\documentclass[letterpaper, 10 pt, conference]{ieeeconf}

\usepackage{mathtools}
\usepackage{graphicx, subcaption}
\graphicspath{{img/}}
\usepackage{epstopdf}
\usepackage{pgf}
\usepackage{calc}
\newcommand\inputpgf[2]{{
\let\pgfimageWithoutPath\pgfimage
\renewcommand{\pgfimage}[2][]{\pgfimageWithoutPath[##1]{#1/##2}}
\input{#1/#2}
}}

\usepackage{todonotes}
\usepackage{soul}
\usepackage{etoolbox}
\usepackage{amsmath}
\usepackage{amssymb}
\usepackage{bm}
\usepackage[export]{adjustbox}
\usepackage{tabularx}
\newcolumntype{Y}{>{\centering\arraybackslash}X} %
\usepackage{tikz}
\usetikzlibrary{bayesnet}
\newlength{\state} %
\usepackage{url}
\usepackage{array}
\usepackage{textcomp}
\usepackage{xspace}
\usepackage{booktabs}
\usepackage{tablefootnote}
\usepackage{cite}
\usepackage{color}
\newcommand{\blue}[1]{{\color{blue}#1}}

\newcommand{\TODO}[1]{
}
\usepackage{hyperref}
\hypersetup{
    colorlinks,
    linkcolor={red!50!black},
    citecolor={blue!50!black},
    urlcolor={blue!80!black}
}
\usepackage[capitalize]{cleveref}
\renewcommand{\Cref}[1]{\cref{#1}} %
\newcommand{\hide}[1]{}
\newcommand{\ie}{\emph{i.e.~}}

\newcommand{\bdmath}{\begin{dmath}}%
\newcommand{\edmath}{\end{dmath}}
\newcommand{\beq}{\begin{equation}}
\newcommand{\eeq}{\end{equation}}
\newcommand{\bdm}{\begin{displaymath}}
\newcommand{\edm}{\end{displaymath}}
\newcommand{\bea}{\begin{eqnarray}}
\newcommand{\eea}{\end{eqnarray}}
\newcommand{\beal}{\beq \begin{array}{ll}}
\newcommand{\eeal}{\end{array} \eeq}
\newcommand{\beas}{\begin{eqnarray*}}
\newcommand{\eeas}{\end{eqnarray*}}
\newcommand{\ba}{\begin{array}} %
\newcommand{\ea}{\end{array}}

\newcommand{\bit}{\begin{itemize}}
\newcommand{\eit}{\end{itemize}}
\newcommand{\ben}{\begin{enumerate}}
\newcommand{\een}{\end{enumerate}}
\newcommand{\bbmat}{\begin{bmatrix}}
\newcommand{\ebmat}{\end{bmatrix}}
\newcommand{\bpmat}{\begin{pmatrix}}
\newcommand{\epmat}{\end{pmatrix}}

\newcommand{\TGV}{$\text{TGV}^2$\xspace}
\newcommand{\NLTGV}{$\text{NLTGV}^2$\xspace}
\newcommand{\FLAME}{FLaME\xspace}

\newcommand{\Euroc}{EuRoC\xspace}

\newcommand{\linkToPdf}[1]{\href{#1}{\blue{(pdf)}}}
\newcommand{\linkToPpt}[1]{\href{#1}{\blue{(ppt)}}}
\newcommand{\linkToCode}[1]{\href{#1}{\blue{(code)}}}
\newcommand{\linkToWeb}[1]{\href{#1}{\blue{(web)}}}
\newcommand{\linkToVideo}[1]{\href{#1}{\blue{(video)}}}
\newcommand{\award}[1]{\xspace} %

\newcommand{\mysubsection}[1]{{\bf#1.}} 
\title{Smooth Mesh Estimation from Depth Data \\ using Non-Smooth Convex Optimization }
\IEEEoverridecommandlockouts
\author{Antoni Rosinol$^{1}$, Luca Carlone$^{1}$
\thanks{$^{1}$A.\,Rosinol and L.\,Carlone are with the Laboratory for
Information \& Decision Systems (LIDS), Massachusetts Institute of Technology, Cambridge, MA, USA,
{\sf \{arosinol,lcarlone\}@mit.edu}}
\thanks{This work was partially funded by ARL DCIST CRA W911NF-17-2-0181, and Lincoln Laboratory's ``Resilient Perception in Degraded Environments'' program.}}%

\begin{document}
\maketitle

\begin{abstract}
    Meshes are commonly used as 3D maps since they encode the topology of the scene while being lightweight.
    Unfortunately, 
    3D meshes are mathematically difficult to handle directly because of their combinatorial and discrete nature.
    Therefore, most approaches generate 3D meshes of a scene after fusing depth data using volumetric or other representations.
    Nevertheless, volumetric fusion remains computationally expensive both in terms of speed and memory.
    In this paper, we leapfrog these intermediate representations 
    and build a 3D mesh directly from a depth map and 
    the sparse landmarks triangulated with visual odometry.
    To this end, we formulate a non-smooth convex optimization problem that we solve using a primal-dual method.
    Our approach generates a smooth and accurate 3D mesh
      that substantially improves the state-of-the-art on direct mesh reconstruction while running in real-time.
\end{abstract}


\TODO{
\section*{Supplementary Material}
\centerline{\href{https://www.mit.edu/~arosinol}{https://www.mit.edu/\texttildelow{arosinol}}}
}

\section{Introduction}
\label{sec:introduction}

Fast, lightweight, and accurate 3D representations of the scene are of utmost importance for computationally constrained robots, such as flying drones,
 or high-stake applications, such as self-driving cars.
Failing to map the scene densely and accurately in a timely manner can lead to collisions,
 which result in damaged equipment or, worst, fatalities.
Being able to densely reconstruct the scene with a camera in real-time also opens a handful of new applications in augmented reality,
 online real-life gaming, or search-and-rescue operations.

Therefore, in this work, we propose a 3D scene reconstruction method that leverages depth data (either from RGB-D or mono/stereo reconstruction) 
and recent advances in the field of convex variational optimization to build a 3D mesh that faithfully represents the scene while working in real-time.

Our work can be used with any available Visual Odometry (VO) pipeline.
In our case, we use a Visual Inertial Odometry (VIO) pipeline,
Kimera \cite{Rosinol20icra-Kimera}, which achieves very accurate pose estimates in real-time by tracking 2D features from frame to frame 
and using inertial information from an Inertial Measuring Unit (IMU).

\begin{figure}[htbp]
  \centering
  \includegraphics[width=0.8\columnwidth]{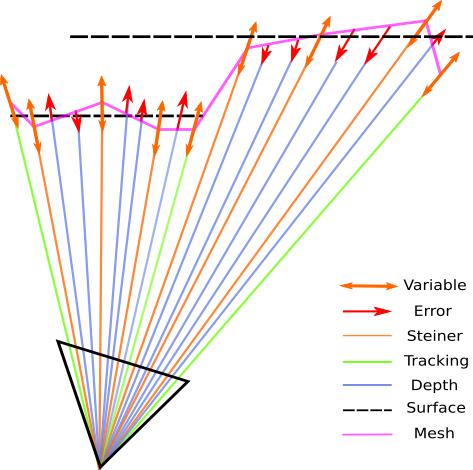}
  \caption{\textit{We propose a 3D mesh reconstruction method from input depth maps (blue rays) coming from RGB-D or dense stereo matching,
   and the sparse landmarks from visual odometry (VO) (green rays).
  We formulate the problem as a variational convex optimization problem which we solve in real-time and to optimality.
  The cost function measures the fitness of the mesh with respect to the depth input (red arrows),
  and the visual odometry landmarks, as well as the smoothness of the mesh.
   The optimization is done over the inverse depth of the vertices of the mesh (orange arrows).}}
  \label{fig:intro}
  \vspace{-1em}
\end{figure}

VO methods typically provide a sparse map of the scene.
In particular, feature-based methods~\cite{Mourikis07icra, Leutenegger15ijrr, Forster17tro, Mur-Artal16arxiv} 
produce a sparse point cloud that is not directly usable for path planning or obstacle avoidance.
In those cases, a denser map is built subsequently, e.g., by using (multi-view) stereo algorithms~\cite{Schoeps17cvpr, Pillai16icra}.
Nevertheless, with current broad availability of depth information from either RGB-D, LIDAR, or stereo cameras, 
it is possible to fuse the depth online using volumetric approaches~\cite{Rosinol21arxiv-Kimera,Newcombe11ismar}.
Unfortunately, such volumetric approaches are memory intensive, and slow if they are not heavily parallelized using powerful GPUs.

Ideally, a map representation should be (i) lightweight to compute and store, and (ii) capable of representing the topology of the environment.
In this work, we use a 3D mesh as our map representation (\Cref{fig:intro}).
A 3D mesh is lightweight, since it can represent piece-wise planar surface with very few parameters (compared to dense pointclouds or volumetric representations),
while providing information about the topology of the scene due to its connectivity information.

Recent approaches have tried to build the mesh over the set of sparse 3D landmarks triangulated by a VO/VIO pipeline \cite{Rosinol19icra-mesh,Greene17flame-iccv}.
Nevertheless, these approaches fail to reconstruct an accurate mesh of the scene,
since they discard most of the information present in the input data.

{\bf Contributions.}
In this paper, we propose \emph{a 3D mesh reconstruction algorithm that summarizes the information present in depth-maps and tracked 3D landmarks from visual odometry.}
In this way, we can build denser and more accurate maps than sparse approaches, while preserving real-time performance.
We experimentally evaluate our approach on the \Euroc{} dataset~\cite{Burri16ijrr-eurocDataset} and the uHumans2 dataset~\cite{Rosinol20rss-dynamicSceneGraphs,Rosinol21arxiv-Kimera},
and compare against competing approaches.
Our evaluation shows that (i) the proposed approach produces a lightweight representation of the environment that captures the geometry of the scene,
(ii)~leveraging variational optimization enables real-time, accurate, and denser reconstructions.

\section{Related Work}
\label{sec:related_work}

While there exist a myriad of representations for 3D mapping,
we focus our related work review on approaches using meshes.
Despite the fact that volumetric approaches are typically used to build a mesh \cite{Curless96siggraph},
 and other authors have tried to mesh intermediate representations such as surfels \cite{Schops19pami-surfelmeshing},
we focus on works that build a 3D mesh directly from input data either in the form of RGB images or RGB-D data.

Meshes were one of the first dense map representations used in the nineties when GPUs were not available.
Early on, two different types of meshes were introduced: 3D tetrahedral meshes and 2D triangular meshes (embedded in 3D).

ProForma~\cite{Fox97cmpb-proforma} was one of the first works to reconstruct a detailed map of the scene 
using a tetrahedral 3D mesh.
Unfortunately, tetrahedral 3D meshes share similar costs and benefits than volumetric approaches,
and hence they were mostly used in small-scale scenes first~\cite{Faugeras90ai,Fox97cmpb-proforma}.
More recent works have used tetrahedral 3D meshes for real-time reconstruction of the scene \cite{Piazza18arxiv-cpu,Lhuillier18cviu-surface},
but the quality and artifacts in the reconstructions remain a problem.
In particular, reconstructing a manifold mesh remains challenging, and many authors have tried to enforce manifoldness
 with different degrees of success \cite{Lhuillier13cviu-manifold,Bodis17cviu-efficient}.
While 3D tetrahedral meshes are getting increasingly more capable, they remain challenging to use, are computationally expensive, and lack robustness.

In this work, we focus instead on building 2D triangular meshes embedded in 3D space.
Turk et al.~\cite{Turk94accgit-zipperedMeshes} proposed Zippered Meshes, an early attempt to fuse depth information into 2D triangular meshes by `sewing' them from frame to frame.
In SLAM, several works have tried to build a 3D mesh directly from triangulation of the features in the image followed by a re-projection in 3D space using the depth estimated
from triangulation.
While very fast in practice, these approaches trade-off quality and accuracy for speed.
Their application has therefore mostly been geared towards fast obstacle avoidance, and local path-planning,
 particularly for resource constrained robots such as drones\cite{Teixeira16iros,Greene17flame-iccv,Rosinol19icra-mesh}, and also on ground-robots~\cite{Putz16,Ruetz19-ovpc}.

One of the major problems of reconstructing a 2D mesh from input data is its lack of robustness to outliers, and smoothness against noise.
By leveraging recent insights in variational smoothing of dense depth maps,
Greene and Roy \cite{Greene17flame-iccv} propose \FLAME, a discrete formulation using a 3D mesh instead of dense depth maps,
thereby achieving substantial computational savings, while reconstructing a smooth 3D mesh.
Since they do not use dense depth maps, readily available from RGB-D, \FLAME suffers from inaccurate reconstructions.
In the same vein, VITAMIN-E\cite{Yokozuka19arxiv-vitamine} shows that tracking a substantially higher amount of 2D features,
 and thereby providing denser depth maps, improves the 3D reconstruction accuracy.
They also used the proposed variational framework from \FLAME,
but they used all tracked features as vertices of the 3D mesh,
rather than summarizing the depth map into a simplified and coherent surface reconstruction.
Finally, \cite{Rosinol19icra-mesh} proposed to smooth the 3D mesh by leveraging the least-squares optimization performed in VO/VIO, and enforcing structural regularities (such as planarity).
While this approach allowed the reconstruction of a multi-frame 3D mesh in real-time, the accuracy remained low due to the sparsity of the features tracked in SLAM.
To cope with the low accuracy of these reconstructions, both VITAMIN-E \cite{Yokozuka19arxiv-vitamine} and Rosinol et al.~\cite{Rosinol19icra-mesh}
fuse the depth into a volumetric representation \cite{Rosinol20icra-Kimera,Rosinol20rss-dynamicSceneGraphs,Rosinol21arxiv-Kimera}.
In a similar turn of events, while Newcombe et al. \cite{Newcombe10cvpr} and Lovegrove \cite{Lovegrove12icl-parametric} initially attempted to build 3D meshes from sparse features,
they concluded that using dense all-pixel methods with volumetric maps leads to far superior accuracy\cite{Newcombe11iccv}, despite the heavy use of GPUs.

In this work, we maintain the speed and low computational budget of early approaches, despite the use of a GPU, by re-using the feature tracks of the visual pipeline,
but we use a dense depth map, from {RGB-D} or stereo, to build an improved and smoothed 3D mesh that summarizes all the depth information.
The resulting 3D mesh best fits the dense depth-maps in an optimal sense, and results in an accurate and lightweight representation of the scene. 
To conclude our review, we note that, more recently, non-euclidean deep learning approaches have started to work with 3D meshes 
and are becoming increasingly popular\cite{Milano20neurips-PDMeshNet,Hanocka19acm-meshcnn,Bloesch19iccv-learning,Feng21arxiv-mesh}.
Unfortunately, these approaches remain slow and require powerful GPUs for computation.

\begin{figure*}[t]
  \centering
  \includegraphics[width=\columnwidth]{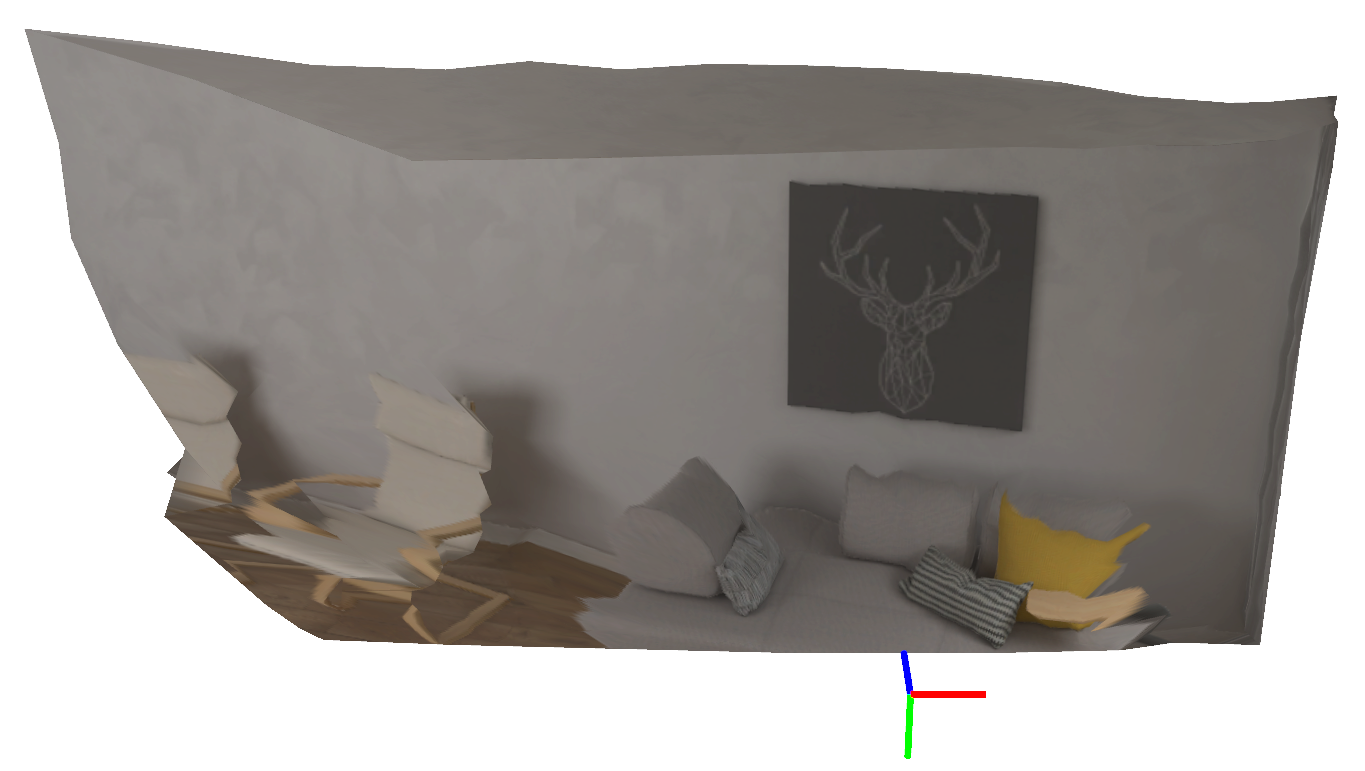}
  \includegraphics[width=\columnwidth]{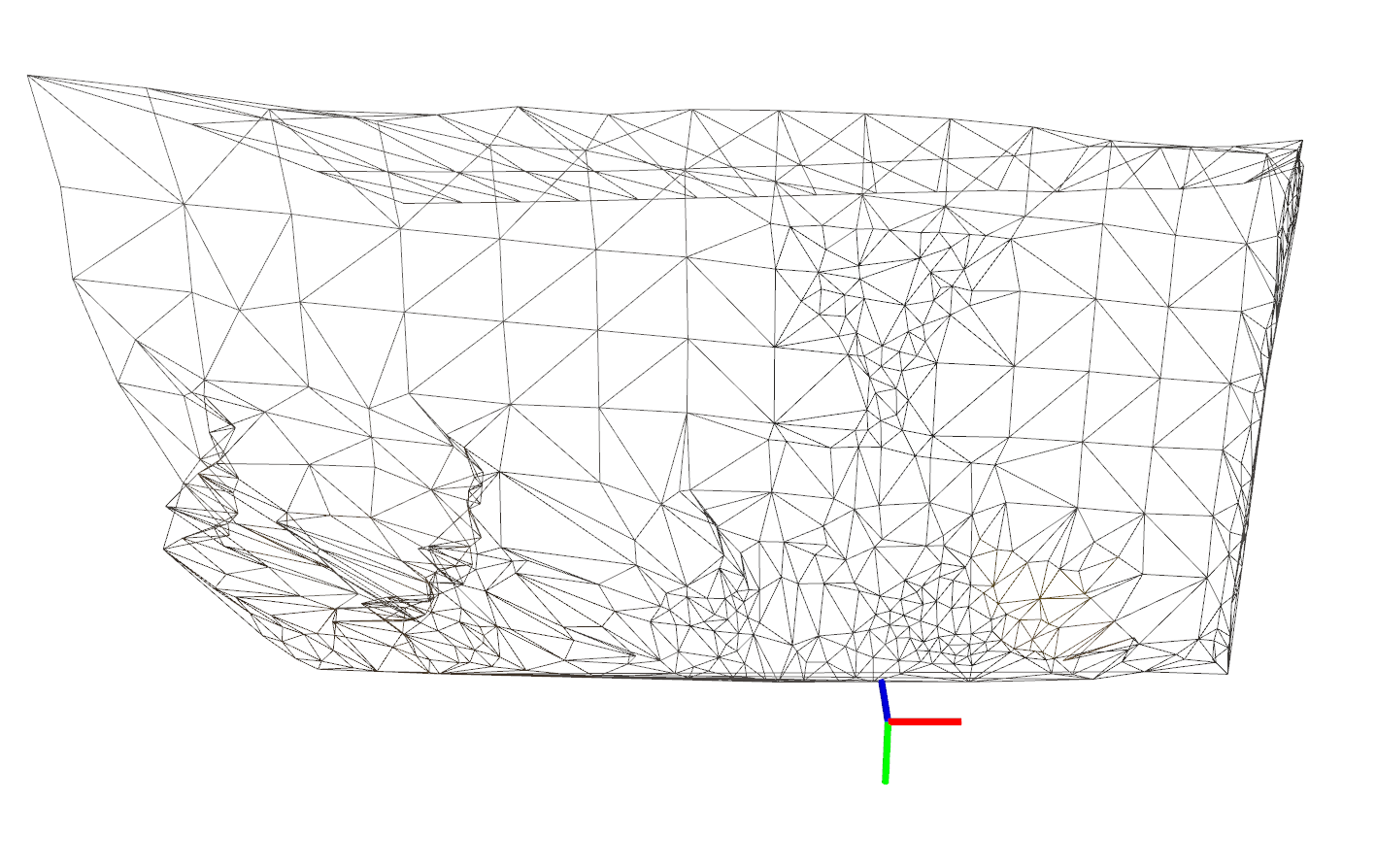}
  \caption{\textit{(Left) Textured 3D mesh reconstruction from the presented approach of a planar surface in the uHumans2 dataset\cite{Rosinol21arxiv-Kimera}.
   (Right) Wireframe of the mesh, showing both interests points (around the painting), and Steiner points attached to the wall.
    We add Steiner points every 50 pixels, forming a squared grid-like pattern.
  }}
  \label{fig:view_uh2}
\end{figure*}

\section{Mesh Optimization: Depth-Map Fusion}

\subsection{Notation}

We follow a similar notation than the one used in \FLAME \cite{Greene17flame-iccv},
 since we use the same mathematical machinery.
We define by $\mathcal{V}$ the set of vertices of the 3D mesh,
and by $\mathcal{E}$ the set of edges of the 3D mesh.
For each edge, $e = (i,j)\in \mathcal{E},$
we denote by $v^{i}, v^{j} \in \mathcal{V}$ the associated vertices.
The edges are directed from $i$ to $j$,
but the sense is arbitrary and can be ignored.
For each vertex, $v \in \mathcal{V}$,
we assign a smoothed inverse depth value that we denote $v_{\xi}$,
and an auxiliary variable $\mathbf{w} \in \mathbb{R}^{2}$ that represents the 2D projection of a 3D normal,
such that $v_{\mathbf{w}}:=$ $(v_{w_{1}}, v_{w_{2}})$.
We encapsulate in $v_{\mathbf{x}} := (v_{\xi}, v_{\mathbf{w}})$ all the (primal) variables associated to a vertex $v$.
We also denote by $\bold{v}_{\xi}$ the inverse depth of all the vertices of the mesh stacked into a column vector.
$\langle\cdot, \cdot\rangle$ is the inner product between two vectors.

\subsection{Fitting the Data}

It is well-known from the computer graphics literature \cite[Section~13.5]{Foley96}
that the inverse depth map $\mathbf{b}$ depends linearly on the inverse depths $\bold{v}_{\xi}$ of the vertices of the mesh:
$$\mathbf{b}=A\bold{v}_{\xi}$$
where the map $A$ encodes the barycentric coordinates of each pixel with respect to the triangle it belongs to.
Here, $\mathbf{b}$ is a column vector with as many entries as there are pixels in the depth image,
 while $\bold{v}_{\xi}$ has as many entries as there are vertices in the mesh.
Therefore, given a 2D mesh (in image space) with fixed topology, we can fit the inverse depths of the vertices such that the mesh minimizes the fit to a given depth-map in a least-squares fashion.
In other words, the loss function for mesh reconstruction could be:
$$\mathcal{L}_{\text{depth}}(\bold{v}_{\xi}) = \|A\bold{v}_{\xi} - \mathbf{b}\|_2^2$$
where $\mathbf{b}$ is now the measured inverse depth, and $A\bold{v}_{\xi}$ is the estimated inverse depth from the mesh.
The optimal value of $\bold{v}_{\xi}$ that solves $\min_{\bold{v}_{\xi}}\mathcal{L}_{\text{depth}}$ is simply given by the normal equations.
Nevertheless, we are not just interested in fitting a 3D mesh to the depth map,
 but rather in summarizing the depth map into a smooth 3D surface.
Moreover, we expect our depth map to potentially contain outlier measurements.
Therefore, we want to formulate our loss function such that it is robust to outliers, and we want to penalize solutions that are not smooth surfaces.

On the one hand, to reduce the influence of outliers, we can make use of the $\ell_1$ norm instead.
In addition, we also want to quantify the difference between the estimated inverse depth $v_\xi$ and the inverse depth $v_z$ given by the visual odometry pipeline, which we denote by $\mathcal{L}_\text{tracking}$.
In summary, our data loss $\mathcal{L}_\text{data}$ consists of two terms,
 $\mathcal{L}_\text{depth}$ and $\mathcal{L}_\text{tracking},$ defined as:
\begin{equation}
\begin{aligned}
  \mathcal{L}_\text{depth}(\bold{v}_{\xi}) &= \sum_{d \in \mathcal{D}} |\mathbf{a}_d \bold{v}_{\xi} - b_d| \\
  \mathcal{L}_\text{tracking}(\bold{v}_{\xi}) &= \sum_{v \in \mathcal{V}}\left|v_{\xi}-v_{z}\right|, 
\end{aligned}
\end{equation}
where $\mathcal{V}$ is the set of vertices of the mesh and $\mathcal{D}$ is the set of depth measurements.
$\mathbf{a}_d, b_d$ are the $d^{\text{th}}$ row of $A$ and $\mathbf{b}$ associated to the $d^{\text{th}}$ depth measurement in $\mathcal{D}$.
While \cite{Teixeira16iros,Greene17flame-iccv,Rosinol19icra-mesh} only used the sparse features from VO tracking to estimate the mesh (using Delaunay triangulation),
 we further add Steiner points \cite[Section~14.1]{De97}, including the set of sparse features tracked by the VO.
A Steiner point is an extra mesh vertex that is added to the optimization problem to create a better fit than it would be possible using only the tracked features from VO alone.
On the other hand, to promote smooth 3D meshes, we can penalize deviations from piecewise affine solutions.
We will quantify this desire by using the $\mathcal{L}_{smooth}$ term below.

We formulate this multi-objective optimization via scalarization,
which introduces the free parameter $\lambda > 0$ balancing the two objectives:
$$\mathcal{L} = \mathcal{L}_{\text{smooth}}+\lambda \mathcal{L}_{\text{data}},$$
where $\mathcal{L}_{\text{data}} = \mathcal{L}_{\text{depth}} + \mathcal{L}_{\text{tracking}}$.
We could also balance the contribution of $\mathcal{L}_{\text{depth}}$ and $\mathcal{L}_{\text{tracking}}$ with another free parameter,
 but we avoid this for simplicity.

\subsection{Smoothing the Mesh}

There are many possible choices for $\mathcal{L}_{smooth}$,
but one that is particularly interesting
 is the second-order Total Generalized Variation (\TGV) functional \cite{Bredies10siam-TGV},
  because it promotes piecewise affine solutions.
A non-local extension was proposed by \cite{Ranftl14eccv-NLTGV}
 that allows using additional priors into the regularization term (\NLTGV).
Pinies et al.\cite{Pinies15icra}
showed that piecewise planarity in image space,
afforded by \NLTGV , translates to piecewise planarity in 3D space, when optimizing over the inverse depth.
Leveraging this insight, Greene and Roy proposed \FLAME \cite{Greene17flame-iccv}
which discretizes \NLTGV for 3D meshes (instead of dense inverse depth maps),
thereby achieving substantial computational savings.
Similarly to \FLAME \cite{Greene17flame-iccv}, we will make use of the discrete \NLTGV to smooth our 3D mesh, which we now explain for 
completeness.

First, we need to introduce a key fact \cite{Pinies15icra}: 
if two pixels $v_{\mathbf{u}}^i$ and $v_{\mathbf{u}}^j$ belong to the same planar surface in 
3D, their inverse depth values $v^i_\xi$ and $v^j_\xi$ are constrained by the following relation:
$$ v_{\xi}^{i}-v_{\xi}^{j}-\left\langle v_{\mathbf{w}}^{i}, v_{\mathbf{u}}^{i}-v_{\mathbf{u}}^{j}\right\rangle = 0,$$
where $v^i_\mathbf{w}$ is the normal vector of the surface at vertex $v^i.$

The \NLTGV smoothing term tries to penalize solutions that deviate from this relationship.
Hence, the \NLTGV term over the mesh is given by:
\begin{equation}
  \label{eq:nltgv}
  \mathrm{NLTGV}^{2}(\xi) \approx \sum_{e \in \mathcal{E}_{k}} \| D_{e}\left(v_{\mathbf{x}}^{i}, v_{\mathbf{x}}^{j}\right)||_{1}
\end{equation}

Here $D_{e}$ is a linear operator that acts on the vertices 
$v_{\mathbf{x}}^{i}$ and $v_{\mathbf{x}}^{j}$ corresponding to edge $e$,
where we recall that $v_\mathbf{x}:= (v_\xi, v_\mathbf{w})$:
$$
{D}_{e}\left(v_{\mathbf{x}}^{i}, v_{\mathbf{x}}^{j}\right)=\left[\begin{array}{c}
e_{\alpha}\left(v_{\xi}^{i}-v_{\xi}^{j}-\left\langle v_{\mathbf{w}}^{i}, v_{\mathbf{u}}^{i}-v_{\mathbf{u}}^{j}\right\rangle\right) \\
e_{\beta}\left(v_{w_{1}}^{i}-v_{w_{1}}^{j}\right) \\
e_{\beta}\left(v_{w_{2}}^{i}-v_{w_{2}}^{j}\right)
\end{array}\right],
$$
where $v_{\mathbf{u}}^{i}$ and $v_{\mathbf{u}}^{j}$ are the 2D pixel projections of the 3D vertices $v^i$ and $v^j$ of the mesh.
The auxiliary variables $v^i_\mathbf{w}$, $v^j_\mathbf{w}$ are related to the 2D projection of the 3D normal of the plane associated to vertices $v^i$ and $v^j$.
Intuitively, the first row of $\| {D}_{e}\left(v_{\mathbf{x}}^{i}, v_{\mathbf{x}}^{j}\right)||_{1}$
 penalizes the total deviation of the vertices from the expected plane going through the vertices,
 while the second and third row 
 penalize having different normals for adjacent vertices.
Weights $e_{\alpha}, e_{\beta} \geq 0$ are assigned to each edge, and control the influence of the smoothness term.
Setting one of these weights to $0$ would cancel the enforcement of planarity between vertices of the mesh.
As in \cite{Greene17flame-iccv}, we set the edge weight
$e_{\alpha}=1 / \|v_{\mathbf{u}}^{i}-v_{\mathbf{u}}^{j}\|_{2},$ \ie inversely proportional to the edge length in pixels, and $e_{\beta}=1$.
It will be useful in the remaining to express the operator ${D}_e$ as a matrix, in particular, we have:
\begin{equation}
\label{eq:edge_operator}  
\begin{aligned}
&{D}_{e}= \\
&\left[\begin{matrix}
e_\alpha & e_\alpha (v_{u_1}^{j}-v_{u_1}^{i}) & e_\alpha (v_{u_2}^{j}-v_{u_2}^{i}) & -e_\alpha & 0        & 0 \\
0        & e_\beta                            & 0                                  & 0         & -e_\beta & 0 \\
0        & 0                                  & e_\beta                            & 0         & 0        & -e_\beta
\end{matrix}\right]
\end{aligned}
\end{equation}
The operator ${D}_{e}$ is applied on the per-edge vector of primal variables, namely: 
$[v_{\mathbf{x}}^i, v_{\mathbf{x}}^j] = [v_{\xi}^i, v_{w_1}^i, v_{w_2}^i, v_{\xi}^j, v_{w_1}^j, v_{w_2}^j].$

\subsection{Primer on Primal-Dual Optimization}

The resulting loss function combining the data terms and \NLTGV smoothness term is:
\begin{equation}
  \label{eq:optimization_problem}
  \begin{split}
    \mathcal{L}(v_\mathbf{x}) &= \mathcal{L}_{\text{smooth}}(v_\mathbf{x})+\lambda \mathcal{L}_{\text{data}}(v_\mathbf{x}) \\
    & = \sum_{e \in \mathcal{E}} \| {D}_{e}\left(v_{\mathbf{x}}^{i}, v_{\mathbf{x}}^{j}\right)||_{1} \\
    & \quad + \lambda \left[ \sum_{v \in \mathcal{V}}|v_{\xi}-v_{z}| + \sum_{d \in \mathcal{D}} |\bold{a}_d \bold{v}_\xi - b_d| \right]\\
  \end{split}
\end{equation}

\Cref{eq:optimization_problem} is a non-smooth convex optimization problem.
The lack of smoothness is due to the use of non-differentiable $\ell_1$ norms.
Fortunately, this type of problems can be solved using first-order primal-dual optimization \cite{Chambolle11jmiv-primalDual}.

Let us first briefly introduce the primal-dual optimization proposed by Chambolle and Pock \cite{Chambolle11jmiv-primalDual}.
The problem considered is the nonlinear primal problem:
\begin{equation}
\label{eq:nonlinear_optimization}
\min _{x \in X} F(K x)+G(x)
\end{equation}
where the map $K : X \rightarrow Y$ is a continuous linear operator, and $G: X \rightarrow[0,+\infty)$
 and $F: Y \rightarrow [0,+\infty)$ are proper, convex, lower semi-continuous functions.

This nonlinear problem can be formulated as the following saddle-point problem \cite{Chambolle11jmiv-primalDual}:
\begin{equation}
  \label{eq:primal_dual_optimization}
  \min _{x \in X} \max _{y \in Y}\langle K x, y\rangle+G(x)-F^{*}(y)
\end{equation}
where $F^{*}$ is the convex conjugate of $F$.
From now on, we refer to variable $x$ as the primal, and $y$ as the dual.
Chambolle and Pock proposed to solve this saddle-point problem by performing descent steps on the primal, and ascent steps on the dual.
In particular, for each iteration $n \geq 0$, the algorithm updates $x^{n}, y^{n}, \bar{x}^{n}$ as follows:
\begin{equation}
\label{eq:primal_dual_iter}
\left\{\begin{aligned}
y^{n+1}&=\left(I+\sigma \partial F^{*}\right)^{-1}\left(y^{n}+\sigma K \bar{x}^{n}\right)  &\text{(Dual ascent)}\\
x^{n+1}&=(I+\tau \partial G)^{-1}\left(x^{n}-\tau K^{*} y^{n+1}\right) &\text{(Primal descent)}\\
\bar{x}^{n+1}&=x^{n+1}+\theta\left(x^{n+1}-x^{n}\right) &
\end{aligned}\right.
\end{equation}
where $\sigma, \tau>0$ are the dual and primal steps, and $\theta \in[0,1]$.
$K^*$ is the adjoint of $K$.
$(I+\tau \partial F^*)^{-1}$ and $(I+\tau \partial G)^{-1}$ are the resolvent operators, 
defined as:
$$
x=(I+\tau \partial G)^{-1}(y)=\arg \min _{x}\left\{\frac{\|x-y\|^{2}}{2 \tau}+G(x)\right\}
$$
The main assumption of \cite{Chambolle11jmiv-primalDual} is that the resolvent operators for $F$ and $G$ are easy to compute.

As noted by Greene and Roy \cite{Greene17flame-iccv},
the summation over the edges $\mathcal{E}$ and the summation over the vertices $\mathcal{V}$
in \Cref{eq:optimization_problem} can be encoded in $F({K}\mathbf{x})$ and $G(\mathbf{x}),$ respectively, in \Cref{eq:nonlinear_optimization}.
It remains to be seen where does the third term $|\bold{a}_d \bold{v}_{\xi} - b_d|$ fit.

\subsection{Finding the Optimal Mesh}

\begin{figure}[htbp]
  \centering
  \includegraphics[width=\columnwidth]{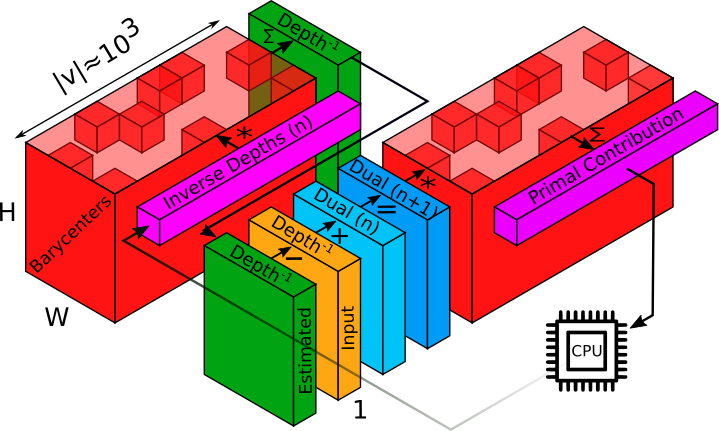}
  \caption{\textit{Graphical representation of the computation required to evaluate the primal contribution in \cref{eq:summary} due to the $H\times W$ dual variable $\mathbf{p}^{n+1}$ (dark blue).
  Starting from the CPU (bottom right), we forward the current inverse depths $v_\mathbf{\xi}^{n}$ of the mesh to the GPU (pink column).
  Then, we rasterize an interpolated inverse depth (using GPU) for the whole image using the barycentric coordinates of each pixel (red, sparse).
  We substract to the interpolated depth (green) the measured depth (yellow),
  and add the previous dual value $\mathbf{p}^{n}$ (clear blue).
  The result is the next dual values (dark blue) $\mathbf{p}^{n+1}$.
  The primal contribution (dark pink) is then calculated by element-wise multiplication with the barycentric coordinates (red, sparse), and sent back to the CPU for a primal update.
  The process is repeated until convergence.
  }}
  \label{fig:gpu_implementation}
  \vspace{-1em}
\end{figure}

On the one hand, we could consider the term $|\bold{a}_d \bold{v}_{\xi} - b_d|$ to be part of $G(\mathbf{x})$, but then we need to find its proximal operator\footnote{The proximal operator is the resolvent of the subdifferential operator.}.
Unfortunately, there is no known closed-form proximal operator for the $\ell_1$ norm of an affine function \cite{Beck17siam}.
On the other hand, we can consider it part of $F({K\mathbf{x}})$.
In this case, we need to dualize the function with respect to $G(\mathbf{x})$.
Unfortunately, doing so will introduce as many dual variables as pixels in the image,
 which ideally we would like to avoid for computational reasons.
We proceed hence with the second choice, and implement a fast GPU-based solver (\cref{fig:gpu_implementation}).

 \TODO{
Alternatively, we can `precondition' our problem by multiplying $A\bold{v}_{\xi} -\mathbf{b}$ with $A^T$, thereby reducing the number of dual variables to the number of vertices in the mesh.
An important caveat of this approach is that $A^TA$ is now,
 potentially, a dense matrix instead of a relatively sparse one as the original $A$ matrix was (three non-zero terms per row, namely the barycentric coordinates).
Also, despite $A$'s sparsity,
it remains a very large matrix,
with $\approx 10^3$ columns and $307200$ rows for a VGA image ($640\times480$).
With $3$ non-zero elements per row, we need to perform $\approx 1M$ operations to generate $A^TA$.
It also remains to be seen if there is a better pre-conditioning factor to this problem,
but this is an active field of research.
}

\TODO{@LUCA I went the DUALS way, bcs not sure I can prove anything good about $A^TA$}
\TODO{PRECONDITION TEXT:
Let us define $\tilde{A} := A^TA$ and $\tilde{\mathbf{b}}:=A^T\mathbf{b}$.
$\tilde{A}$ can be ill-conditioned if a face of the mesh has little measurements (or none) while the others have many...}
\TODO{DUALS TEXT:
Remove tildes... $\tilde{A} := A$ ....and $\tilde{\mathbf{b}}:=\mathbf{b}$.
}

The dual representation of $\|{A}\bold{v}_{\xi}-{\mathbf{b}}\|_1$ is given by (Appendix~\ref{proof:a}):
$$\max_{\mathbf{p}}\langle {A}\bold{v}_{\xi}-{\mathbf{b}}, \mathbf{p}\rangle-\delta_{P}(\mathbf{p}),$$
where $\delta_{P}(\bold{p})$ is the convex conjugate of the $\ell_1$ norm,
and corresponds to the indicator function of the set $P = \{\mathbf{p}: \|\mathbf{p}\|_\infty \leq 1\}$.
Together with the dual representation of the \NLTGV term \cite{Greene17flame-iccv}:
$$ \left\|{D}_{e}\left(v_{\mathbf{x}}^{i}, v_{\mathbf{x}}^{j}\right)\right\|_{1}=
\max _{e_{\mathbf{q}}}\left\langle{D}_{e}\left(v_{\mathbf{x}}^{i}, v_{\mathbf{x}}^{j}\right), e_{\mathbf{q}}\right\rangle-\delta_{Q}\left(e_{\mathbf{q}}\right) $$
where each $e_\mathbf{q} \in \mathbb{R}^{3}$ corresponds to a dual variable for each edge,
and $\delta_{Q}(e_{\mathbf{q}})$ is also the convex conjugate of the $\ell_1$ norm.
We have the primal-dual formulation of our original problem:
\begin{equation}
  \label{eq:primal_dual_formulation}
  \begin{aligned}
    \min_{v_{\mathbf{x}}} \max_{e_{\mathbf{q}}, \mathbf{p}}
    & \sum_{e \in \mathcal{E}_{k}}\left\langle{D}_{e}\left(v_{\mathbf{x}}^{i}, v_{\mathbf{x}}^{j}\right), e_{\mathbf{q}}\right\rangle-\delta_{Q}\left(e_{\mathbf{q}}\right) \\
    & +\lambda \left[\sum_{v \in \mathcal{V}_{k}}\left|v_{\xi}-v_{z}\right| 
      + \langle {A}\bold{v}_{\xi}-{\mathbf{b}}, \mathbf{p}\rangle-\delta_{P}(\mathbf{p})\right]\\
  \end{aligned}
\end{equation}

Let us now formulate this primal-dual optimization problem under the formalism of \cref{eq:primal_dual_optimization}.
In particular, \cref{eq:primal_dual_formulation} can be cast as \cref{eq:primal_dual_optimization} by defining $F^{*}$ and $G$ as follows:

\begin{equation*}
  \begin{aligned}
    F^{*}(e_\mathbf{q}, \mathbf{p})&= \sum_{e \in \mathcal{E}_{k}}\delta_{Q}(e_\mathbf{q}) + \lambda \langle {\mathbf{b}}, \mathbf{p}\rangle + \delta_{P}(\mathbf{p}) ,  \\
    G(\mathbf{v}_\xi)&=\lambda\sum_{v \in \mathcal{V}_{k}}|v_\xi-v_z|.
  \end{aligned}
\end{equation*}

We now detail the resolvent operators $\left(I+\sigma \partial F^{*}\right)^{-1}$ and $(I+\tau \partial G)^{-1}$.
Since $F^{*}$ is the sum of separable functions, we can first find the resolvent for $q$ and then $p$.
For the indicator function of a convex set $\delta_Q(q)$, the resolvent operator reduces to point-wise projectors onto unit $\ell_{\infty}$ balls.
Similarly, for the sum of an indicator function of a convex set $\delta_P(\mathbf{p})$ and the linear map $\lambda\langle\mathbf{b}, \mathbf{p}\rangle$, 
the resolvent operator remains a point-wise projector onto the unit $\ell_{\infty}$ ball, where the argument is offset by ${\lambda\mathbf{b}}$.
$$
(q, p)=\left(I+\sigma \partial F^{*}\right)^{-1}(\tilde{q}, \tilde{p}) \Leftrightarrow
\left\{
  \begin{aligned}
    q_{i}&=\frac{\tilde{q}_{i}}{\max \left(1,\left|\tilde{q}_{i}\right|\right)} \\
    p_{j}&=\frac{\tilde{p_{j}}-\lambda{b}_{j}}{\max \left(1,\left|\tilde{p_{j}}-\lambda{b}_{j}\right|\right)}
  \end{aligned}\right.
$$

Similarly, the resolvent operator for $G(\mathbf{v}_\xi)$ is given by the pointwise shrinkage operations:
$$
\begin{aligned}{}
v_\xi&=(I+\tau \partial G)^{-1}(\tilde{v_\xi})
\Leftrightarrow  \\
v_{\xi}&=
\left\{\begin{aligned}{}
\tilde{v_\xi}-\tau \lambda & \text { if } & \tilde{v_\xi}-v_z & > & \tau \lambda \\
\tilde{v_\xi}+\tau \lambda & \text { if } & \tilde{v_\xi}-v_z & < & -\tau \lambda \\
v_z & \text { if } & \left|\tilde{v_\xi}-v_z\right| & \leq & \tau \lambda
\end{aligned}\right.
\end{aligned}
$$

We now detail the primal descent and dual ascent steps.
First, we need to define the linear operator $K$ for the dual ascent.
The dual ascent over $\mathbf{q, p}$ is given by:
\begin{equation}
  \label{eq:dual_ascent}
  \left(
  \begin{aligned}
  &\mathbf{q}^{n+1} \\
  &\mathbf{p}^{n+1}
  \end{aligned}
  \right)
  =
  \left(I+\sigma \partial F^{*}\right)^{-1}
  \left(
  \begin{aligned}
    &\mathbf{q}^{n}+\sigma D\mathbf{v}_{\mathbf{x}} \\
    &\mathbf{p}^{n}+\sigma\lambda{A}\mathbf{v}_\xi
  \end{aligned}
  \right)
\end{equation}
where we use $D$ to represent the linear operator that stacks all of the ${D}_e$ operators acting on a per-edge basis.
$\sigma>0$ is the dual step size.

\TODO{$v_x$ and $v_\xi$ should be bold, no?}
\TODO{A bit sketchy that $v_{\mathbf{x}}$ has in it $v_\xi$ and yet we have different dual variables that affect $v_\xi$...
Maybe we should just drop the tracking term...}

Let us now define the adjoint $K^*$ for the primal descent over $v_\xi$.
Since we have formulated our linear operators in terms of matrices ${D}_e$ and ${A}$,
the adjoints are simply the transpose of these matrices: ${D}_e^* = {D}_e^T$ and ${A}^* = {A}^T$.
We now further decompose ${D}_e^*$ to make the subsequent primal descent iteration explicit.
${D}_e^*$ operates on the incoming edges $\mathcal{N}_{in}(v)$
and outgoing edges $\mathcal{N}_{out}(v)$ to a vertex $v$, as noted by \cite{Greene17flame-iccv}, and therefore can be decomposed:
$$
\begin{aligned}
{D}_{e}^{*}\left(e_{\mathbf{q}}\right)= D_e^T e_\mathbf{q} = &\left[\begin{array}{c}
e_{\alpha} e_{q_{1}} \\
e_{\alpha}\left(v_{u_1}^{j}-v_{u_1}^{i}\right) e_{q_{1}}+e_{\beta} e_{q_{2}} \\
e_{\alpha}\left(v_{u_2}^{j}-v_{u_2}^{i}\right) e_{q_{1}}+e_{\beta} e_{q_{3}} \\
-e_{\alpha} e_{q_{1}} \\
-e_{\beta} e_{q_{2}} \\
-e_{\beta} e_{q_{3}}
\end{array}\right] \\
&=\left[\begin{array}{c}
{D}_{i n}^{*}\left(e_{\mathbf{q}}\right) \\
{D}_{o u t}^{*}\left(e_{\mathbf{q}}\right)
\end{array}\right]
\end{aligned}
$$
where ${D}_{i n}^{*}$ and ${D}_{out}^{*}$ partition $D_e^*$ into the first and last three rows, respectively.
Furthermore, we need to make explicit the primal updates on the inverse depths, since the adjoint $A^*$ only affects these.
Therefore, we use that:
$$
{D}_{e}^{*}\left(e_{\mathbf{q}}\right)
=
\left[\begin{array}{c}
{D}_{i n}^{*}\left(e_{\mathbf{q}}\right) \\
{D}_{o u t}^{*}\left(e_{\mathbf{q}}\right)
\end{array}\right]
=
\left[\begin{array}{c}
{D}_{in_\xi}^{*}\left(e_{\mathbf{q}}\right) \\
{D}_{in_{\mathbf{w}}}^{*}\left(e_{\mathbf{q}}\right) \\
{D}_{out_\xi}^{*}\left(e_{\mathbf{q}}\right) \\
{D}_{out_{\mathbf{w}}}^{*}\left(e_{\mathbf{q}}\right)
\end{array}\right]
$$

The descent equations for each $v_{\xi}$ are then given by:
\begin{equation}
  \label{eq:primal_ascent_1}
  \begin{aligned}
    v_{\xi}^{n+1}= (I + \tau \partial G)^{-1} \bigg( v_{\xi}^n &-\tau \sum_{e \in \mathcal{N}_{i n}(v)} {D}_{in_\xi}^{*} \left(e_{\mathbf{q}}^{n+1}\right) \\
    &-\tau  \sum_{e \in \mathcal{N}_{out}(v)} {D}_{out_\xi}^{*}\left(e_{\mathbf{q}}^{n+1}\right)\\
    &-\tau\lambda[{A}^* \mathbf{p}^{n+1}]_v \bigg),
  \end{aligned}
\end{equation}
where $\tau>0$ is the primal step size, and $[\cdot]_v$ in $[{A}^* \mathbf{p}^{n+1}]_v$ extracts the row corresponding to vertex $v$.
The descent equations for each $v_{\mathbf{w}}$ are given by:
\begin{equation}
  \label{eq:primal_ascent_2}
  \begin{aligned}
  v_{\mathbf{w}}^{n+1}= v_{\mathbf{w}}^{n} - \tau & \sum_{e \in \mathcal{N}_{i n}(v)} {D}_{in_\mathbf{w}}^{*} \left(e_{\mathbf{q}}^{n+1}\right) \\
  -\tau & \sum_{e \in \mathcal{N}_{out}(v)} {D}_{out_\mathbf{w}}^{*}\left(e_{\mathbf{q}}^{n+1}\right).
  \end{aligned}
\end{equation}

With the primal descent \cref{eq:primal_ascent_1,eq:primal_ascent_2} and dual ascent \cref{eq:dual_ascent} in hand, we can now iteratively apply \cref{eq:primal_dual_iter} until convergence.

In summary, our optimization proceeds as follows:
\begin{equation}
\label{eq:summary}
\left\{\begin{aligned}
&\quad\quad\quad\underline{\text{Dual ascent:}}& \\
  \left(
  \begin{aligned}
  &\mathbf{q}^{n+1} \\
  &\mathbf{p}^{n+1}
  \end{aligned}
  \right)
  &=
  \left(I+\sigma \partial F^{*}\right)^{-1}
  \left(
  \begin{aligned}
    &\mathbf{q}^{n}+\sigma D\mathbf{v}_{\mathbf{x}} \\
    &\mathbf{p}^{n}+\sigma\lambda{A}\mathbf{v}_\xi
  \end{aligned}
  \right) \\
&\quad\quad\quad\underline{\text{Primal descent:}}& \\
  \left(
  \begin{aligned}
  &v_{\xi}^{n+1}\\
  &v_{\mathbf{w}}^{n+1}
  \end{aligned}
  \right)
  &=
  \left(I+\sigma \partial G\right)^{-1}
  \left(
  \begin{aligned}
    v_\xi^{n}&-\tau {D}^*_\xi e_\mathbf{q^{n+1}}\dots \\ 
    &\dots-\tau \lambda[{A}^*\mathbf{p}^{n+1}]_v \\
    v_{\mathbf{w}}^{n}&-\tau {D}^*_\mathbf{w} v_\xi
  \end{aligned}
  \right) \\
&\quad\quad\quad\underline{\text{Extra gradient step:}}& \\
\bar{v}^{n+1}_\mathbf{x}&=v_\mathbf{x}^{n+1}+\theta\left(v_\mathbf{x}^{n+1}-v_\mathbf{x}^{n}\right)
\end{aligned}\right.
\end{equation}
and we implement the dual step in the GPU as illustrated in \cref{fig:gpu_implementation}.

\section{Experimental Results}
\label{sec:results}

\TODO{1. Single depth map compression into 3D mesh: (left) raw depth map; (right) meshed depth map}

\TODO{2. TUM experiment with RGB-D data}

\TODO{3. Euroc experiment with stereo dense reconstruction}

We benchmark the proposed approach on real and simulated datasets, and evaluate map estimation accuracy, as well as runtime.

The \Euroc dataset~\cite{Burri16ijrr-eurocDataset} provides visual and inertial data recorded from a micro aerial vehicle flying indoors.
We use the \textit{Vicon Room} (\texttt{V}) data, which is similar to an office room where walls,
floor, and ceiling are visible, as well as other planar surfaces (boxes, stacked mattresses).

We also use the simulated uHumans2 dataset \cite{Rosinol21arxiv-Kimera} to evaluate the accuracy of the mesh reconstruction with RGB-D data.
The uHumans2 dataset is generated using a photorealistic Unity-based simulator, and provides RGB-D data and ground truth of the geometry of the scene.
Using this dataset, we can also provide an ablation study on the effect that the number of Steiner points has on the reconstruction (without being influenced by stereo reconstruction errors).
We will be using the `Apartment` scene in the uHumans2 dataset.

\mysubsection{Compared techniques}
To assess the advantages of our proposed approach, we compare the original \FLAME formulation, and ours, as well as VO pipelines.
We also evaluate with both RGB-D data, using uHumans's dataset, and dense stereo reconstruction, using the \Euroc dataset.

\subsection{Evaluation in \Euroc}
\label{ssec:mapping_quality_euroc}

\begin{figure*}[htbp]
  \centering
  \includegraphics[width=0.5\columnwidth]{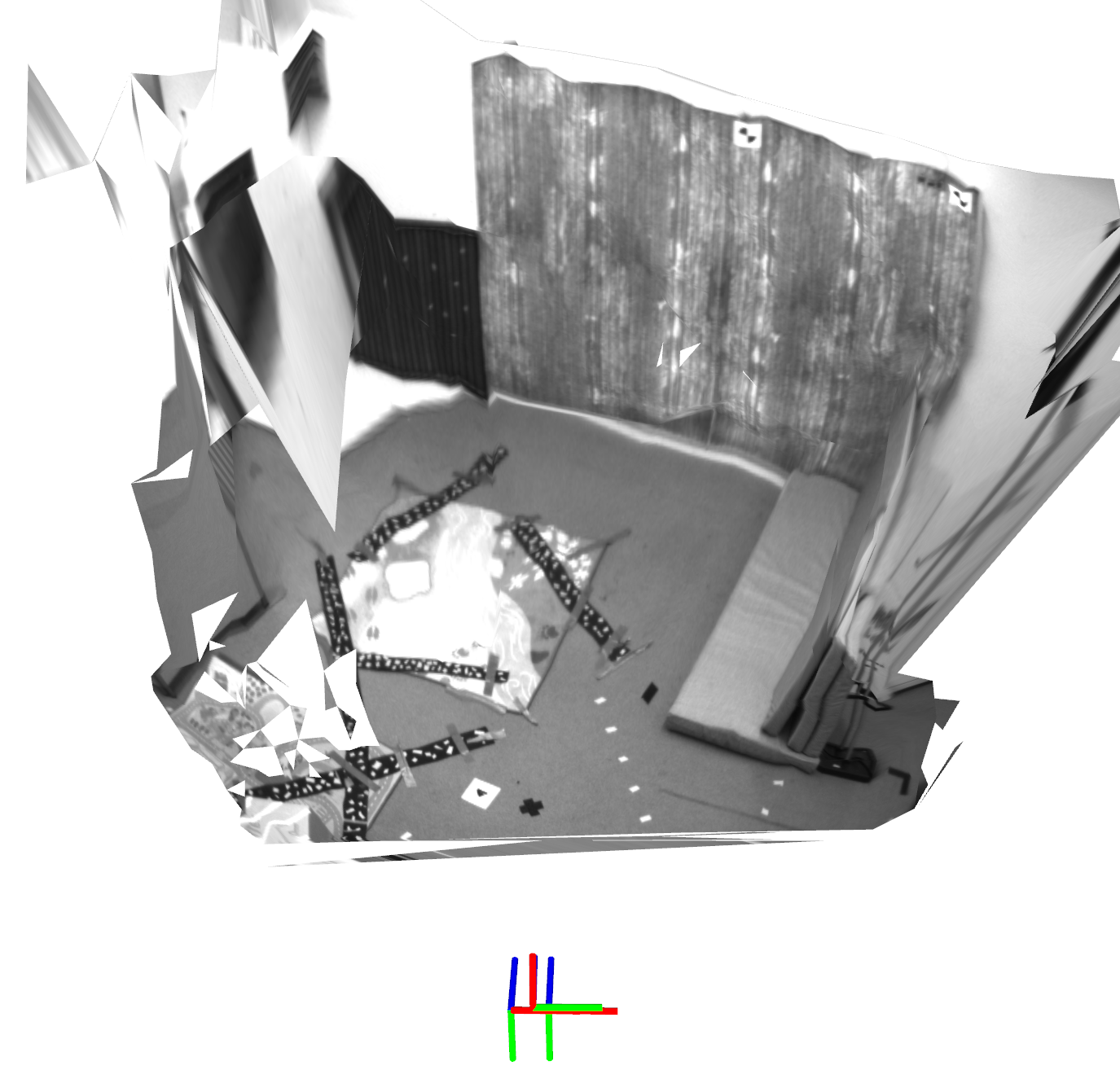}
  \includegraphics[width=0.5\columnwidth]{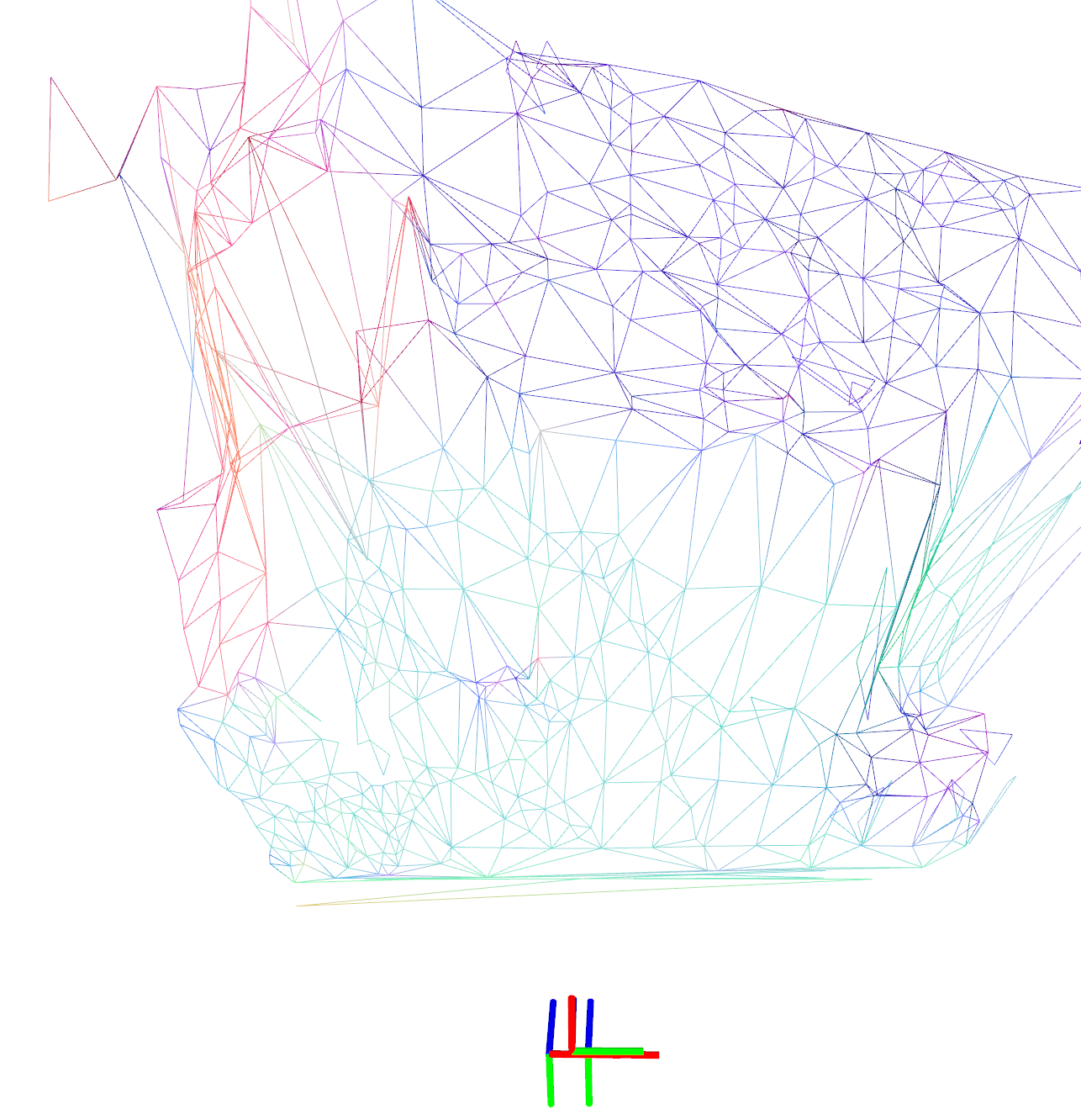}
  \includegraphics[width=0.5\columnwidth]{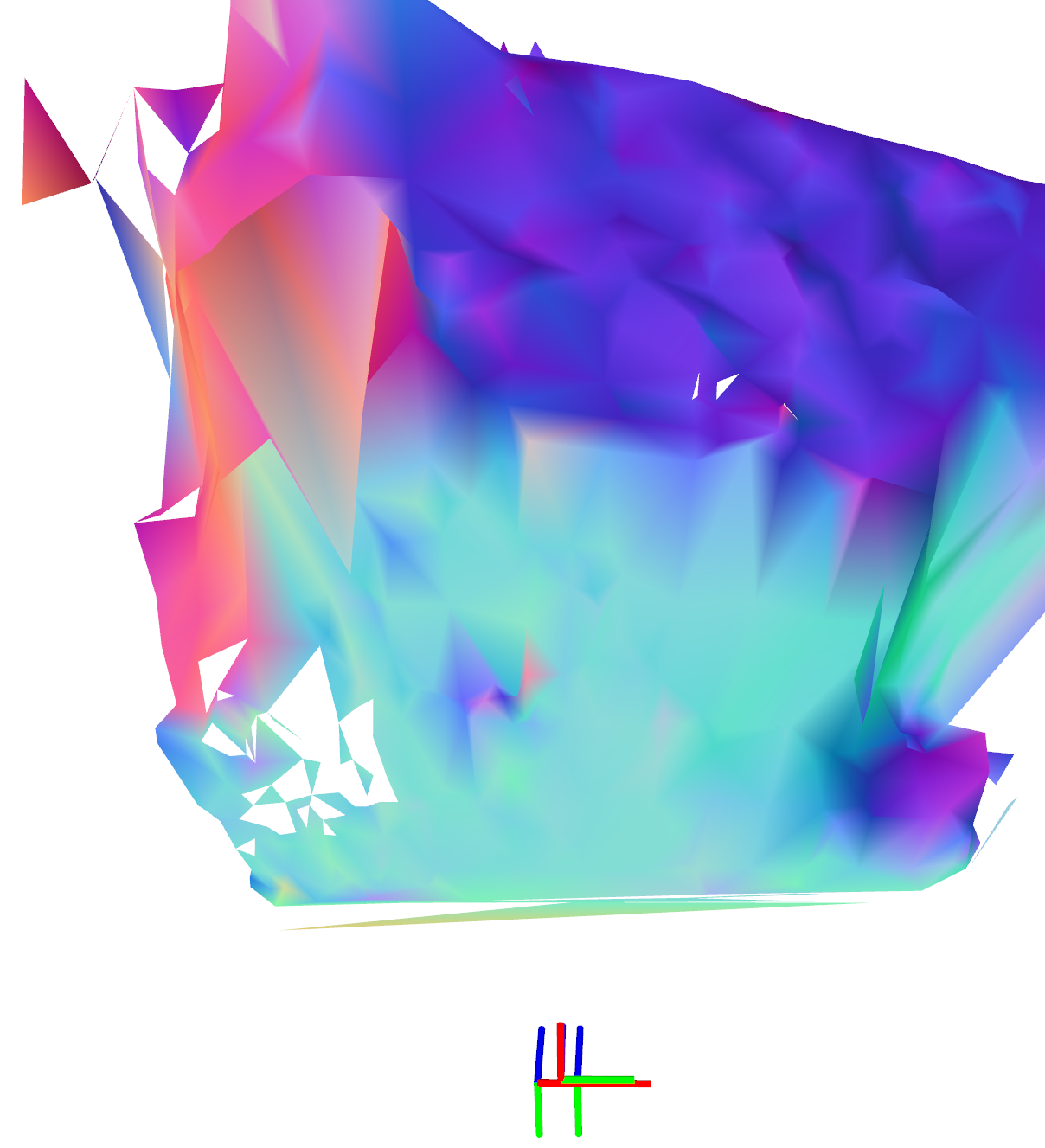}
  \includegraphics[width=0.5\columnwidth]{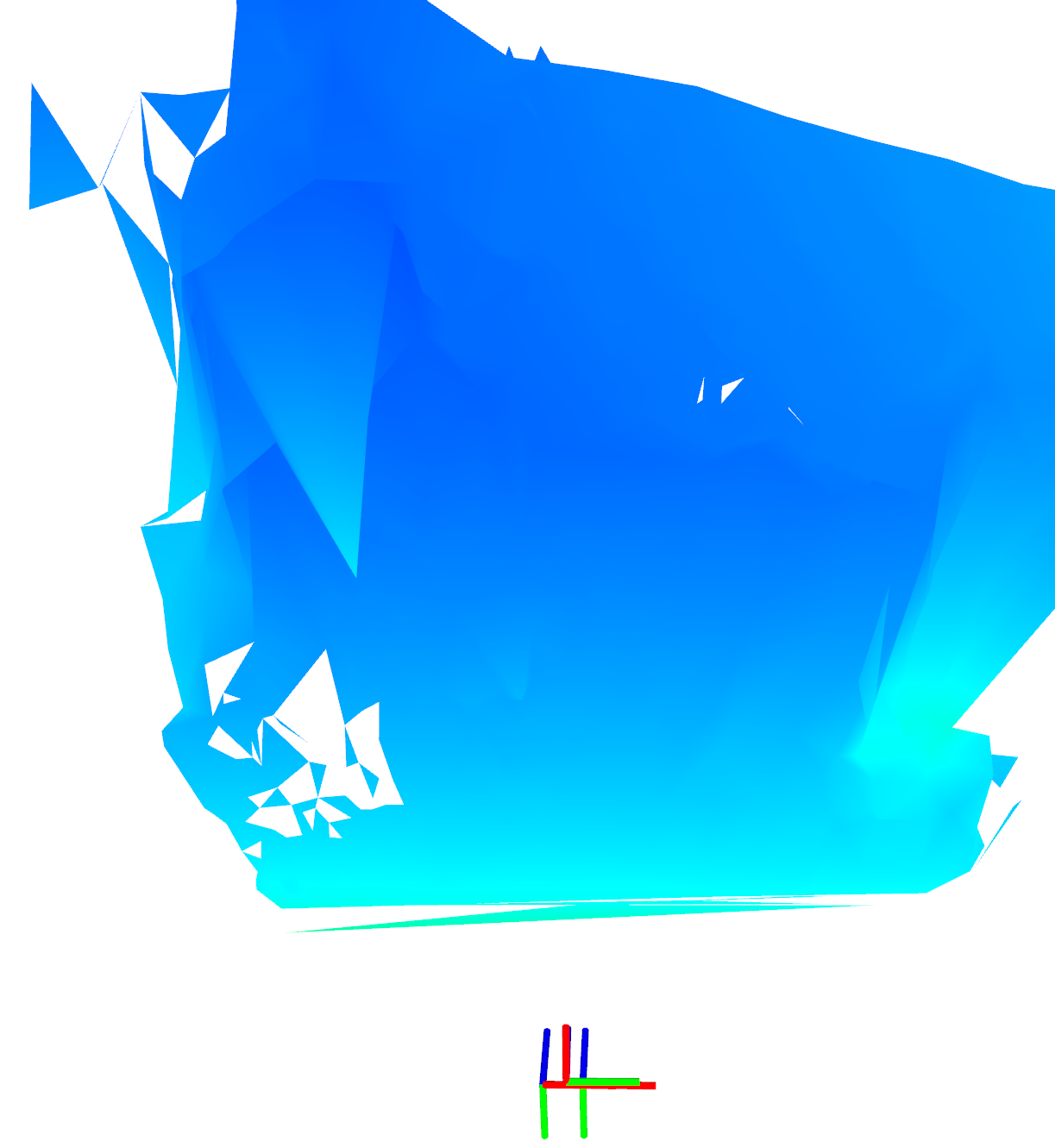}
  \caption{\textit{(First) Textured 3D mesh showing the proposed mesh reconstruction by using dense stereo depth estimation in the \Euroc dataset\cite{Burri16ijrr-eurocDataset}.
   (Second) Wireframe of the mesh. (Third) Normal map showing the smoothness of the 3D mesh. (Fourth) Color-coded depth map, brighter colors correspond to closer depths.}}
  \label{fig:view_euroc}
\end{figure*}

The \Euroc dataset provides ground-truth 3D reconstructions of the scene for the Vicon Room 
(\texttt{V}) datasets.
Therefore, we can directly evaluate the accuracy of the per-frame 3D mesh reconstructions.
Nevertheless, the dataset was recorded with a stereo camera instead of an RGB-D camera.
Hence, we perform dense stereo reconstruction (using OpenCV's block-matching approach) to fit our 3D mesh.

\Cref{tab:mesh_accuracy_euroc} compares our approach with two VO pipelines, LSD-SLAM\cite{Engel14eccv-lsdslam} and MLM\cite{Greene16icra-mlm},
as well as \FLAME\cite{Greene17flame-iccv}, which is the work closest to our approach.
LSD-SLAM is a semi-dense VO pipeline, and as such, it partially reconstructs the scene.
While being semi-dense, LSD-SLAM remains practically unable to reconstruct the scene, staying below $25\%$ in terms of accurate depth estimates.
MLM is built on top of LSD-SLAM and was designed to densify LSD-SLAM's map, as such, it clearly achieves denser reconstructions than its counterpart (approximately $50\%$ denser reconstructions).
\FLAME further densifies the 3D reconstruction by using 3D meshes and a primal-dual formulation like ours.
Their approach clearly improves the MLM estimates (by an average factor of $42\%$), but most importantly, provides topological information with the 3D mesh.
Our approach further uses the depth data coming from RGB-D or stereo, and fuses this information into the 3D mesh, which further densifies the 3D reconstruction,
 while also keeping the topological properties of the scene.
\Cref{fig:view_euroc} shows the quality of the reconstruction in \Euroc.

\begin{table}[htbp!]
  \centering
  \caption{
    \textit{
   Fraction of inverse depths per depth-map 
   within 10 percent of ground-truth 
   for LSD-SLAM\cite{Engel14eccv-lsdslam}, MLM\cite{Greene16icra-mlm},
    \FLAME\cite{Greene17flame-iccv}, and our approach.
   Our approach enables more accurate and denser reconstructions.
   In \textbf{bold} the best result.}}
  \label{tab:mesh_accuracy_euroc}
  \begin{tabularx}{\columnwidth}{l *4{Y}}
    \toprule
    \Euroc & \multicolumn{4}{c}{Accurate Inverse Depth Density [\%]} \\
    \cmidrule(l){2-5}
    Sequence  & LSD & MLM & \FLAME & Ours \\
    \midrule
      V1\_01 & 18.2 & 26.4 & 37.8 & \textbf{41.7} \\
      V1\_02 & 18.3 & 27.9 & 40.5 & \textbf{44.8} \\
      V1\_03 & 11.0 & 17.0 & 27.0 & \textbf{33.1} \\
      V2\_01 & 25.9 & 39.3 & 48.1 & \textbf{52.9} \\
      V2\_02 & 20.5 & 28.9 & 37.2 & \textbf{44.0} \\
      V2\_03 & 11.5 & 19.0 & 28.9 & \textbf{34.8} \\
    \bottomrule
  \end{tabularx}
  \vspace{-1em}
\end{table}

\TODO{We use the ground truth point cloud for dataset \texttt{V1} to assess the quality of the 3D mesh by calculating its \textit{accuracy},
 as defined in \cite{Schoeps2017cvpr}.
To compare the mesh with the ground truth point cloud,
 we compute a point cloud by sampling the mesh with a uniform density of $10^3~\text{points}/m^2$.
We also register the resulting point cloud to the ground truth point cloud.
  In \cref{fig:accuracy_mesh},
   we color-encode each point $r$ on the estimated point cloud with its distance 
   to the closest point in the ground-truth point cloud $\mathcal{G}$ ($d_{r \to \mathcal{G}}$).
}

\subsection{Evaluation in uHumans2 Dataset}
\label{ssec:mapping_quality_uHumans2}

Using the RGB-D images in uHumans2 dataset, we run and evaluate our approach using an increasing number of Steiner points.
\Cref{tab:mesh_accuracy_uh2} shows that increasing the number of Steiner points has a clear positive effect on the quality of the depth estimates.
The improvements are most noticeable when comparing the use of no Steiner points (No Steiner), and adding Steiner points every 100 pixels ($S=100$).
Subsequently, the addition of Steiner points increases the accuracy up to around one Steiner point every 10 pixels.
Note that the computation time increases according to the number of points added,
and beyond $S=50$ our approach might not be fast enough for real-time operations.
\Cref{fig:view_uh2} shows a reconstruction using the $S=50$ configuration.

\begin{table}[htbp!]
  \centering
  \caption{
    \textit{
   Fraction of inverse depths per depth-map 
   within 10 percent of ground-truth 
   for our approach using an increasing number of Steiner points every $S$ pixels.
   The accuracy and density of our reconstructions increase with the number of Steiner points added.
   In \textbf{bold} the best result.}}
  \label{tab:mesh_accuracy_uh2}
  \begin{tabularx}{\columnwidth}{l *4{Y}}
    \toprule
    uHumans2 & \multicolumn{4}{c}{Accurate Inverse Depth Density [\%] (Timing [ms])} \\
    \cmidrule(l){2-5}
    Sequence  & Ours (No Steiner) & Ours ($S=100$) & Ours ($S=50$) & Ours ($S=10$)\\
    \midrule
      Apart\_01 & 43.2 (9.1) & 46.4 (10.9) & 53.8          (12.4) & \textbf{56.7} (41)\\
      Apart\_02 & 42.3 (10.2)& 49.9 (11.9) & \textbf{52.5} (13.4) & 51.2          (49)\\
      Apart\_03 & 43.0 (9.8) & 48.0 (11.8) & 53.0          (13.3) & \textbf{53.9} (43)\\
    \bottomrule
  \end{tabularx}
  \vspace{-1em}
\end{table}

\subsection{Timing}
\label{ssec:timing}

All timings were processed using an Nvidia RTX 2080 Ti GPU on an Intel i9 CPU.
The computation of the dual contribution takes less than $2.1$ms overall (worst case timing recorded, including data transfer).
Nevertheless, since this operation needs to be repeated several times until convergence, we record an average processing time of $13.2$ms.
This includes performing the Delaunay triangulation, uploading depth and 2D mesh coordinates to the GPU,
solving the dual problem, sending the primal contribution to the CPU, optimizing, and repeating the process until convergence (for $6$ iterations).
Ultimately, the bottleneck in terms of computation comes from Kimera-VIO which takes $20$ms per frame,
despite being an optimized VIO pipeline that has its frontend and backend working in parallel.

\section{Conclusion}
\label{sec:conclusions}

By leveraging the triangulated landmarks from visual odometry and the depth from RGB-D data (or stereo),
we present a 3D reconstruction algorithm capable of building a smooth,
 compact, and accurate 3D mesh of the scene.
The proposed algorithm shows that dense depth fusion into a consistent 3D mesh is not only possible, 
but also computationally fast.

Finally, while the results presented are promising, we are not yet fusing 
the 3D mesh into a consistent multi-frame scene representation.
Therefore, a promising line for future work is to use the per-frame 3D mesh to achieve multi-view 3D scene reconstruction.
With the accuracy and speed afforded by the presented approach,
it becomes possible to build a map without relying on intermediate volumetric representations. 
\IEEEtriggeratref{71}
\bibliographystyle{IEEEtran}
\section{Appendix}
\label{proof:a}
Let us show that the following dual representation holds:
$$\|A\boldsymbol{\xi}-\mathbf{b}\|_1 = \max_{\mathbf{p}\in{\mathbb{R}^n}}\langle A\boldsymbol{\xi}-\mathbf{b}, \mathbf{p}\rangle-\delta_{P}(\mathbf{p}),$$
where $\boldsymbol{\xi}, \mathbf{b, p} \in \mathbb{R}^n$, $A \in \mathbb{R}^{n\times n}$,
and
$\delta_{P}(\mathbf{p})$ is the indicator function of the convex set $P = \{\mathbf{p} : \|\mathbf{p}\|_\infty \leq 1\}$.
We assume in the remaining that $A$ is nonsingular.

\subsubsection{Proposition \cite[Section~3.3]{Boyd04book}} 
Since $A$ is nonsingular, the conjugate of $g(\mathbf{x})=$ $f(A\mathbf{x}+\mathbf{b})$ is:
$$ g^{*}(\mathbf{x})=f^{*}\left(A^{-T} \mathbf{x}\right)-b^{T} A^{-T} \mathbf{x},$$
with $\mathbf{dom}~g^{*}=A^{T} \mathbf{dom}~f^{*}$, with $f^*$ the conjugate of $f$.

\subsubsection{Proposition \cite[Section~3.3]{Boyd04book}}
The convex conjugate of the $\ell_1$ norm is the indicator function 
$\delta_Y(\mathbf{y})$ of the convex set $Y = \{\mathbf{y} : \|\mathbf{y}\|_\infty \leq 1\}$.

Using proposition 1 and 2,
we have that the convex conjugate of $g(\mathbf{x})=\|A\mathbf{x}-\mathbf{b}\|_1 = f(A\mathbf{x}-\mathbf{b})$, is: 
$$g^{*}(\mathbf{x})=\delta_{Y}(A^{-T}\mathbf{x}) + \mathbf{b}^{T} A^{-T} x,$$
where $Y = \{A^{-T}\mathbf{x} : \|A^{-T}\mathbf{x}\|_\infty \leq 1\}$, 
and $\mathbf{dom}~g^{*}=A^{T} \mathbf{dom}~f^{*} = \{\mathbf{x} : \|A^{-T}\mathbf{x}\|_\infty \leq 1\}$.
Let us take the convex conjugate of $g^{*}$, by definition, we have:
$$ g^{**}(\mathbf{y}) = \max_{\mathbf{x}} \langle \mathbf{y} , \mathbf{x} \rangle - \left[ \delta_Y(A^{-T}\mathbf{x}) + \mathbf{b}^TA^{-T}\mathbf{x}\right] $$
\TODO{By strict definition, it is $\sup_x$ not $\max_x$, show that $\sup_x$ is in $X$...}
\TODO{also, specify $\mathbf{dom}~g^{**}$ (the domain of the conjugate function consists of $x\in \mathbb{R}^n$ for which the supremum is finite)}
Taking $\mathbf{x}^\prime = A^{-T}\mathbf{x}$, we have:
$$ 
\begin{aligned}
    g^{**}(\mathbf{y}) &= \max_{\mathbf{x}^\prime} \langle \mathbf{y}, A^T\mathbf{x}^\prime \rangle - \left[ \delta_X^\prime(\mathbf{x}^\prime) + \mathbf{b}^T\mathbf{x}^\prime\right] \\
              &= \max_{\mathbf{x}^\prime} \langle A\mathbf{y} - \mathbf{b}, \mathbf{x}^\prime \rangle - \delta_{X^\prime}(\mathbf{x}^\prime) \\
\end{aligned},
$$
where $X^\prime = \{\mathbf{x}^\prime : \|\mathbf{x}^\prime\|_\infty \leq 1\}$

Since $g(\mathbf{x})$ is a proper, lower semi-continuous, and convex function, we have $g^{**} = g$, according to Fenchel-Moreau's theorem.
Therefore, 
$$ 
g^{**}(\boldsymbol{\xi}) = g(\boldsymbol{\xi}) = \|A\boldsymbol{\xi}-\mathbf{b}\|_1 = \max_{\mathbf{p}} \langle A\boldsymbol{\xi} - \mathbf{b}, \mathbf{p} \rangle - \delta_P(\mathbf{p}),
$$
with $P = \{\mathbf{p} : \|\mathbf{p}\|_\infty \leq 1\}$.

\end{document}